# Approximating MAP using Local Search

James D. Park and Adnan Darwiche
Computer Science Department
University of California
Los Angeles, CA 90095
{jd,darwiche}@cs.ucla.edu

## Abstract

MAP is the problem of finding a most probable instantiation of a set of variables in a Bayesian network, given (partial) evidence about the complement of that set. Unlike computing priors, posteriors, and MPE (a special case of MAP), the time and space complexity of MAP is not only exponential in the network treewidth, but also in a larger parameter known as the "constrained" treewidth. In practice, this means that computing MAP can be orders of magnitude more expensive than computing priors, posteriors or MPE. For this reason, MAP computations are generally avoided or approximated by practitioners.

We have investigated the approximation of MAP using local search. The local search method has a space complexity which is exponential only in the network treewidth, as is the complexity of each step in the search process. Our experimental results show that local search provides a very good approximation of MAP, while requiring a small number of search steps. Practically, this means that the average case complexity of local search is often exponential only in treewidth as opposed to the constrained treewidth, making approximating MAP as efficient as other computations.

## 1 Introduction

The task of computing the Maximum a Posterior hypothesis (MAP) is to find the most likely configuration of a set of variables (which we call the MAP variables) given (partial) evidence about the complement of that set (the non-MAP variables).

One specialization of MAP, which has received a lot of attention, is the Most Probable Explanation (MPE) [15]. MPE is the problem of finding the most likely configuration of a set of variables given a particular instantiation of the complement of that set. The primary reason for this attention to MPE is that it seems to be a much simpler problem than its MAP generalization.

Unfortunately, MPE is not always suitable for the task at hand. For example, in system diagnosis, where the health of each component is represented using a variable, one is interested in finding the most likely configuration of health variables only—the likely input and output values for each component are not of interest. Additionally, the projection of an MPE solution on these health variables is not necessarily a most likely configuration. Nor is the configuration which results from choosing the most likely state of each variable separately.

Computing MAP seems to be significantly more difficult than computing priors, posteriors or MPE. All of these problems are NP-Hard, including their approximations [1, 3], but the computational resources needed to solve MAP using state-of-the-art algorithms are much greater than those needed to compute MPE, for example. Suppose that we decide to solve MAP and MPE using a variable elimination algorithm [16, 8]. Although we can use any elimination order to compute MPE, we can only use a subset of these orders to compute MAP. Specifically, for an elimination algorithm to be sound for MAP, it requires that we eliminate the non-MAP variables first. This reduces the space of elimination orders, possibly throwing out the most efficient orders from consideration. As an example, consider the network in Figure 1, which admits 6 different elimination orders. Any of these orders can be used to solve MPE. To compute MAP of variables B,C, however, only two of these orders can be used and the width of each is 2. Note that we could use an order of width 1 for computing MPE in this case.

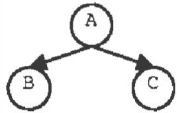

| Order | Width | MPE Order | MAP Order |
|-------|-------|-----------|-----------|
| ABC | 2 | X | X |
| ACB | 2 | X | X |
| BAC | 1 | X | |
| BCA | 1 | X | |
| CAB | 1 | X | |
| CBA | 1 | X | |

Figure 1: A simple network, its possible elimination orders, the widths, and whether or not each order can be used for MPE, and MAP(B,C). Requiring that A be eliminated before B and C forces the width of the elimination order used for MAP(B,C) to be 2, while an order of width 1 can be used for MPE.

The complexity of a variable elimination algorithm is exponential in the *(induced) width* of the used elimination order.[1] Hence, the increase in such a width when computing MAP can be critical: it may simply make a particular network inaccessible to variable elimination algorithms when computing MAP, even though it is accessible when computing MPE.

In order to assess the magnitude of increase in width caused by restricting elimination orders, we generated 1000 Bayesian networks randomly as given in Appendix A and then computed the constrained and unconstrained elimination orders for these networks using the min-fill heuristic [12, 9]. For constrained orders, all non-MAP variables were eliminated first. Each network had 100 nodes and the set of MAP variables consisted of 10-25 root nodes. We measured the minimum, maximum, average and weighted average width for the two classes of orders. The average was computed as $\sum_{i=1}^{k} w_i/k$. Since the complexity is exponential in the width, a weighted sum gives a better representation of the average complexity. It was computed as $log_2(\sum_{i=1}^{k} 2^{w_i}/k)$. Figure 2 summarizes the results.

In many cases, the constrained width was much larger than the unconstrained width, often making the MAP problem unreasonably expensive, even when the MPE problem could be solved exactly with reasonable resources. For example, the weighted average width increased from about 13 to about about 27 due to MAP constraints. That is, even though the largest table constructed by variable elimination has about $2^{14}$ entries when computing MPE, the algorithm needs to construct a table with about $2^{28}$ entries when computing MAP.

The additional resources needed to solve MAP are not only a property of variable elimination algorithms, but are also shared by other algorithms, such as clustering [13, 10, 9] and conditioning [4]. There is definitely a gap between our ability to solve MAP and MPE problems, which is best witnessed by the lack of support for MAP algorithms in existing commercial tools for Bayesian network inference.

In this paper, we propose and investigate a method for approximating MAP using local search. The local search method has a space complexity which is exponential only in the network treewidth, as is the complexity of each step in the search process. Our experimental results show that local search provides a very good approximation of MAP, while requiring a small number of search steps. Practically, this means that the average case complexity of local search is often exponential only in treewidth as opposed to the constrained treewidth, making MAP computations as efficient as other computations.[2]

## 2 Approximating MAP using Local Search

Given a Bayesian network $\mathcal{B}$ which induces a probability distribution $Pr$, and given a set of MAP variables **S**, the goal of a MAP algorithm is to compute an instantiation **s** that maximizes $Pr(\mathbf{s} \mid \mathbf{e})$ for some evidence **e**.[3]

Since computing MAP is often intractable, approximation techniques are needed. A common approximation technique is to compute an MPE and then project the result on the MAP variables. That is, if **S'** is the complement of variables **S** ∪ **E**, we compute an instantiation **s, s'** that maximizes $Pr(\mathbf{s}, \mathbf{s'} \mid \mathbf{e})$ and then return **s**. Another approximation is to compute posterior marginals for MAP variables, $Pr(S \mid \mathbf{e}), S \in \mathbf{S}$, and then choose the most likely state $s$ of each vari-able given **e**. In [7] genetic algoritms were applied to approximate the best k configurations of the MAP variables (this problem is known as partial abduction). We investigate in this paper a different approximation technique based on local search, which works as follows:

1. Start from an initial guess **s** at the solution.

2. Iteratively try to improve the solution by moving to a better neighbor $\mathbf{s'}$: $Pr(\mathbf{s'} \mid \mathbf{e}) > Pr(\mathbf{s} \mid \mathbf{e})$, or equivalently $Pr(\mathbf{s'}, \mathbf{e}) > Pr(\mathbf{s}, \mathbf{e})$.

---

[1] The width of an elimination order with respect to a network is defined as the size of the maximal clique -1 in a jointree constructed based on the elimination order. It can also be equivalently defined as the number of variables -1 in the largest table constructed when running variable elimination using the order.

[2] The network treewidth is defined as the width of its best elimination order. The constrained treewidth is defined as the width of its best constrained elimination orders; hence, is defined with respect to a set of MAP variables.

[3] We are using the standard notation: variables are denoted by upper-case letters ($A$) and their values by lower-case letters ($a$). Sets of variables are denoted by bold-face upper-case letters (**A**) and their instantiations are denoted by bold-face lower-case letters (**a**).

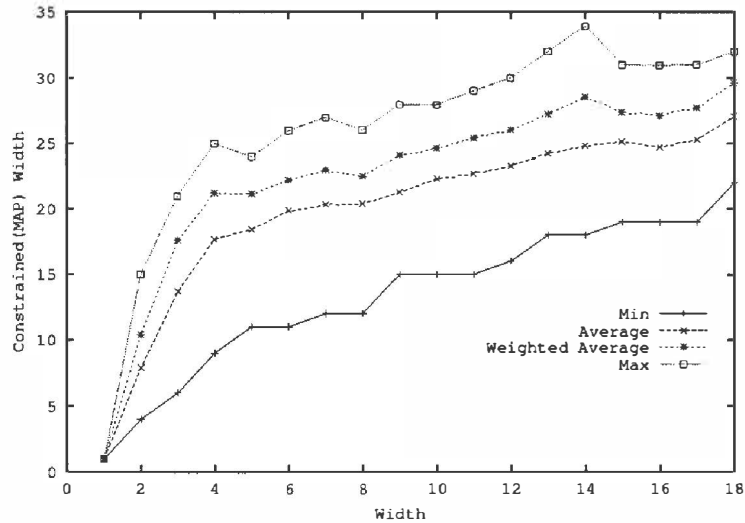

Figure 2: The minimum, maximum, average and weighted average widths (both constrained and unconstrained). Notice that the constrained width can grow to be unmanageable even for networks with small unconstrained width.

A neighbor of instantiation **s** is defined as an instantiation which results from changing the value of a single variable $X$ in **s**. If the new value of $X$ is $x$, we will denote the resulting neighbor by $\mathbf{s} - X, x$. In order to perform local search efficiently, we need to compute the scores for all of the neighbors $\mathbf{s} - X, x$ efficiently. That is, we need to compute $Pr(\mathbf{s} - X, x, \mathbf{e})$ for each $X \in \mathbf{S}$ and each of its values $x$ not in **s**. If variables have binary values, we will have $|\mathbf{S}|$ neighbors in this case.

Local search has been proposed as a method for approximating MPE [11, 14]. For MPE, the MAP variables **S** contain all variables which are not in **E** (the evidence variables). Therefore, the score of a neighbor, $Pr(\mathbf{s}-X, x, \mathbf{e})$, can be computed easily since $\mathbf{s}-X, x, \mathbf{e}$ is a complete instantiation. In fact, given that we have computed $Pr(\mathbf{s}, \mathbf{e})$, the score $Pr(\mathbf{s} - X, x, \mathbf{e})$ can be computed in constant time.[4]

---
[4]This assumes that none of entries in the CPTs are 0.

Unlike MPE, computing the score of a neighbor, $Pr(\mathbf{s}-X, x, \mathbf{e})$, in MAP requires a global computation since $\mathbf{s} - X, x, \mathbf{e}$ may not be a complete instantiation. One of the main observations underlying our approach, however, is that the score $Pr(\mathbf{s} - X, x, \mathbf{e})$ can be computed in $O(n \exp(w))$ time and space where $n$ is the number of network variables and $w$ is the width of a given elimination order (*we can use any elimination order for this purpose, no need for any constraints*). In fact, we can even do better than this by computing the scores of all neighbors $Pr(\mathbf{s} - X, x, \mathbf{e})$ (that is, for all $X \in \mathbf{S}$ and every value $x$ of $X$) in $O(n \exp(w))$ time and space. There are a couple of ways to do this. We can use a modification of the technique of fast retraction in jointrees, which requires working with a special kind of a jointree [2]. An alternative, more direct approach is to use differential inference [5].

According to this approach, the probability distribution of a Bayesian network can be represented as a multivariate polynomial $P(\lambda_x, \ldots)$, in which we have a variable $\lambda_x$ for each value $x$ of each network variable. Variables $\lambda_x$ are called evidence indicators as we can use them to capture evidence: The probabil-

---
If there are 0 entries in the CPTs, it may take time linear in the number of network variables to compute the score. $Pr(\mathbf{s}, \mathbf{e})$ is the product of the single entry of each CPT that is compatible with $\mathbf{s}, \mathbf{e}$. When changing the state of variable $X$ from $x$ to $x'$, the only values in the product that change are those from the CPTs of $X$ and its children. If none of the CPT entries are 0, $Pr(\mathbf{s} - X, x', \mathbf{e})$ can be computed by dividing $Pr(\mathbf{s}, \mathbf{e})$ by the old and multiplying by the new entry for the CPTs for $X$ and its children. This can be done in constant time if the number of children is bounded by a constant.

ity of some evidence e can be obtained by evaluating the polynomial $P$ while setting each indicator $\lambda_x$ to 1 if $x$ is consistent with e and to 0 otherwise. The value of the polynomial under these indicator settings is denoted by $P(\mathbf{e})$. As is shown in [5]:

$$Pr(\mathbf{s} - X, x, \mathbf{e}) = \partial P(\mathbf{s}, \mathbf{e})/\lambda_x.$$

Moreover, we can compute the above partial derivatives for all $\lambda_x$ in only $O(n\exp(w))$ time and space.[5] This means that if we have an elimination order of width $w$ for the given Bayesian network, then we can perform each search step in $O(n\exp(w))$ time and space. As we shall see later, it takes a small number of search steps to obtain a good MAP solution. Hence, the overall runtime is often $O(n\exp(w))$ too. Therefore, we can solve MAP in time and space which are exponential in the unconstrained width instead of the constrained one, which is typically much larger.

The local search method proposed in this section differs from the local search methods used for MPE in that the unconstrained width must be small enough so that a search step can be performed relatively efficiently. It is pointless to use this method to approximate MPE since in the time to take one step, the MPE could be computed exactly. *This method is applicable when the unconstrained width is reasonable but the constrained width is not (see Figure 2).*

## 3 Description of the Methods Used

### 3.1 Search Methods

We tested two common local search methods, *hill climbing* with random restart and *taboo search*. They differ mainly in how they proceed once a peak (local maximum) is reached.

Hill climbing with random restart proceeds by repeatedly changing the the state of the variable that creates the maximum probability change. When a peak is reached, a series of random moves are taken to get to a new start location. Figure 3 gives the algorithm explicitly.

Another variant of hill climbing we implemented is taboo search. Taboo search is similar to hill climbing except that the next state is chosen as the best state that hasn't been visited recently. Because the number of iterations is relatively small we save all of the previous states so that at each iteration a unique point is chosen. Pseudocode for taboo search appears in Figure 4.

### 3.2 Initialization

The quality of the solution returned by a local search routine depends to a large extent on which part of the search space it is given to explore. We implemented several algorithms to compare the solution quality with different initialization schemes. Suppose that $n$ is the number of network variables, $w$ is the width of a given elimination order, and $m$ is the number of MAP variables.

1. *Random initialization (Rand)*. For each MAP variable, we select a value uniformly from its set of states. This method takes $O(m)$ time.

2. *MPE based initialization (MPE)*. We compute the MPE solution given the evidence. Then, for each MAP variable, we set its value to the value that the variable takes on in the MPE solution. This method takes $O(n\exp(w))$ time.

3. *Maximum likelihood initialization (ML)*. For each MAP variable $X$, we set its value to the instance $x$ that maximizes $Pr(x \mid \mathbf{e})$. This method takes $O(n\exp(w))$ time.

4. *Sequential intialization (Seq)*. This method considers the MAP variables $X_1,\ldots,X_m$, choosing each time a variable $X_i$ that has the highest probability $Pr(x_i \mid \mathbf{e}, \mathbf{y})$ for one of its values $x_i$, where $\mathbf{y}$ is the instantiation of MAP variables considered so far. This method takes $O(mn\exp(w))$ time.

## 4 Experimental Results

Two search methods (*Hill* and *Taboo*) and four initialization methods (*Rand, MPE, ML, Seq*) lead to 8 possible algorithms. Each of the initialization methods can also be viewed as an approximation algorithm since one can simply return the computed initialization. This leads to a total of 12 different algorithms. We experimentally evaluated and compared 11 of these

---

[5]The view of a network distribution as a multivariate polynomial $P(\lambda_x,\ldots)$ is what motivated our investigation of local search methods. Specifically, the probability of instantiation s corresponds to the value of polynomial $P$ under a particular indicators setting ($\lambda_x = 1$ if $x$ is consistent with s and $\lambda_x = 0$ otherwise.) This allows us to view the computation of MAP as an optimization problem where we are looking for the values of indicators $\lambda_x$ (instantiation s) that maximize the value of polynomial $P$. A natural way for addressing this problem is to use gradient descent search, especially that computing the gradient $\partial P/\partial \lambda_x$ can be done efficiently. Interestingly enough, the derivative $\partial P(\mathbf{s})/\partial \lambda_x$ is nothing but the probability of current instantiation s after having changed a single variable $X$ to $x$, $Pr(\mathbf{s} - X, x)$. Our initial approach was to implement a standard gradient descent method, where we take a small step in the direction of the gradient. But we then realized that the presented (simpler) approach works quite well, so we opted for it instead.

---

Given: Probability distribution $Pr$, evidence $e$, MAP variables $S$.
Compute: An instantiation $s$ which (approximately) maximizes $Pr(s \mid e)$.

Initialize current state $s$.
$s_{best} = s$
Repeat many times:
    Compute the score $Pr(s - X, x, e)$ for each neighbor $s - X, x$.
    If no neighbor has a higher score that the score for $s$ then
        Repeat for several times
            $s = s'$ where $s'$ is a randomly selected neighbor of $s$.
    Else
        $s = s'$ where $s'$ is the neighbor with the highest score.
    If $Pr(s, e) > Pr(s_{best}, e)$ then
        $s_{best} = s$
Return $s_{best}$

---

Figure 3: Hill climbing with random restart. Notice that when the algorithm reaches a peak, it performs a random walk to get to the next state.

---

Given: Probability distribution $Pr$, evidence $e$, MAP variables $S$.
Compute: An instantiation $s$ which (approximately) maximizes $Pr(s \mid e)$.

Initialize current state $s$.
$s_{best} = s$
Repeat many times
    Add $s$ to *visited*
    Compute the score $Pr(s - X, x, e)$ for each neighbor $s - X, x$.
    $s = s'$ where $s'$ is a neighbor with the highest score not in *visited*.
    If no such neighbor exists (this rarely occurs)
        Repeat for several times
            $s = s'$ where $s'$ is a randomly selected neighbor of $s$.
    If $Pr(s, e) > Pr(s_{best}, e)$ then
        $s_{best} = s$
Return $s_{best}$

---

Figure 4: Taboo search. Notice that the action taken is to choose the best neighbor that hasn't been visited. This leads to moves that decrease the score after a peak is discovered.

---

algorithms, leaving out the algorithm corresponding to random initialization.

To test the quality of various algorithms, we generated random network structures using two generation methods (see Appendix A). For each structure, we quantified the CPTs for different bias coefficients from 0 (deterministic except the roots), to .5 (values chosen uniformly) so we could evaluate the influence of CPT quantification on the solution quality. Each network consisted of 100 variables, with some of the root variables chosen as the MAP variables. If there were more than 25 root variables, we randomly selected 25 of them for the MAP variables. Otherwise we used all of the root variables. We chose root nodes for MAP variables because typically some subset of the root nodes are the variables of interest in diagnostic applications. Evidence was set by instantiating leaf nodes. Care was taken to insure that the instantiation had a non zero probability. Each algorithm was allowed 150 network evaluations.[6] We computed the true MAP and compared it to the solutions found by each algorithm. Additionally, we measured the number of network evaluations needed to find the solution each algorithm subsequently returned, and the number of peaks discovered before that solution was discovered.

We generated 1000 random network structures for each of the two structural generation methods. For each random structure generated, and each quantification method, we quantified the network, computed the exact MAP, and applied each of the approximation algorithms. Figures 5 and 6 show the solution quality of each of the methods by reporting the fraction of networks that were solved correctly; that is, the approximate answer had the same value as the exact answer.

---

[6] An evaluation takes $O(n \exp(w))$ time and space, where $n$ is the number of network variables and $w$ is the width of given elimination order.

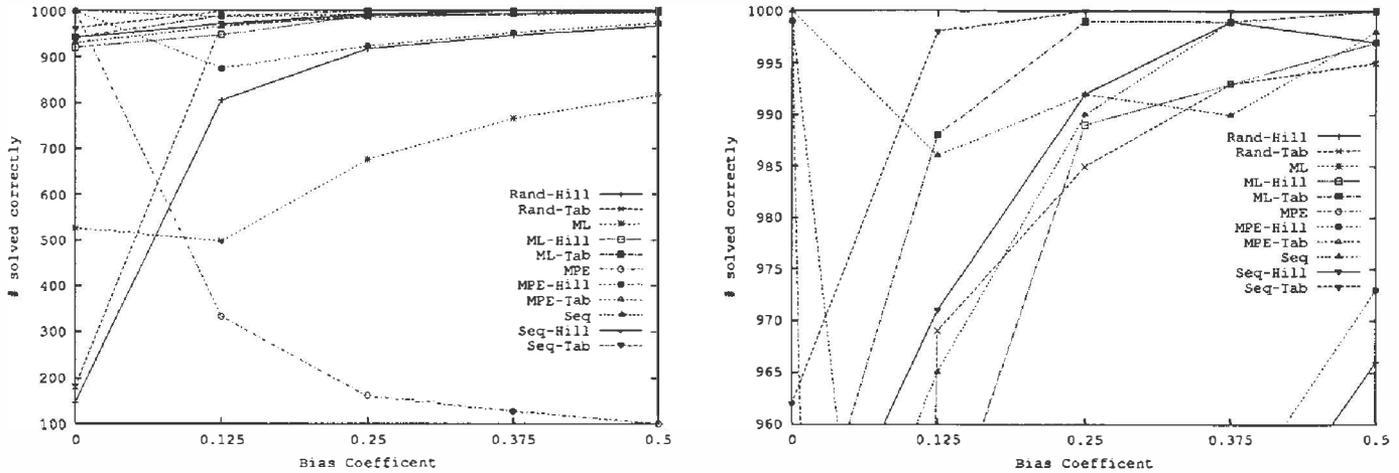

Figure 5: The solution quality of the various search and initialization methods for the first random generation method. The y-axis is the number of problems solved correctly out of 1000. The x-coordinate is the bias coefficient used for quantifying the CPTs. The plot on the right is a zoomed view of the one on the left. The corresponding raw data appears in table 1.

Data Set 1 Solution Quality

|  | 0 | .125 | .250 | .375 | .5 |
|---:|---|---|---|---|---|
| Rand-Hill | 147 | 805 | 917 | 946 | 966 |
| Rand-Taboo | 181 | 969 | 985 | 993 | 995 |
| ML | 526 | 497 | 676 | 766 | 817 |
| ML-Hill | 920 | 947 | 989 | 993 | 997 |
| ML-Taboo | 942 | 988 | 999 | 999 | **1000** |
| MPE | 999 | 333 | 160 | 127 | 100 |
| MPE-Hill | 999 | 875 | 923 | 952 | 973 |
| MPE-Taboo | **1000** | 986 | 992 | 990 | 998 |
| Seq | 930 | 965 | 990 | 999 | 997 |
| Seq-Hill | 941 | 971 | 992 | 999 | 997 |
| Seq-Taboo | 962 | **998** | **1000** | **1000** | **1000** |

Table 1: The solution quality of each method for the first data set. This data is the same as displayed in figure 5. The number associated with each method and bias is the number of instances solved correctly out of 1000. The best scores for each bias are shown in bold.

Data Set 2 Solution Quality

|  | 0 | .125 | .250 | .375 | .5 |
|---:|---|---|---|---|---|
| Rand-Hill | 20 | 634 | 713 | 799 | 845 |
| Rand-Taboo | 20 | 851 | 907 | 943 | 965 |
| ML | 749 | 453 | 495 | 519 | 514 |
| ML-Hill | 966 | 922 | 947 | 963 | 962 |
| ML-Taboo | 973 | 960 | 986 | 987 | 990 |
| MPE | 858 | 505 | 365 | 275 | 206 |
| MPE-Hill | 961 | 853 | 850 | 874 | 891 |
| MPE-Taboo | 978 | 952 | 962 | 977 | 980 |
| Seq | 988 | 955 | 964 | 985 | 972 |
| Seq-Hill | 988 | 960 | 966 | 986 | 976 |
| Seq-Taboo | **994** | **977** | **990** | **994** | **994** |

Table 2: The solution quality of each method for the second data set. This data is the same as displayed in figure 6. The number associated with each method and bias is the number of instances solved correctly out of 1000. The best scores for each bias are shown in bold.

One can draw a number of observations based on these experiments:

- In each case, taboo search performed slightly better than hill climbing with random restarts.

- The search methods were typically able to perform much better than the initialization alone.

- Even from a random start, the search methods were able to find the optimal solution in the majority of the cases.

- Overall, taboo search with sequential initialization performed the best, but required the most network evaluations.

Table 3 contains some statistics on the number of network evaluations (including those used for initialization) needed to achieve the value that the method finally returned. The mean number of evaluations is quite small for all of the methods. Surprisingly, for the hill climbing methods, the maximum is also quite small. In fact, after analyzing the results we discovered that the hill climbing methods never improved

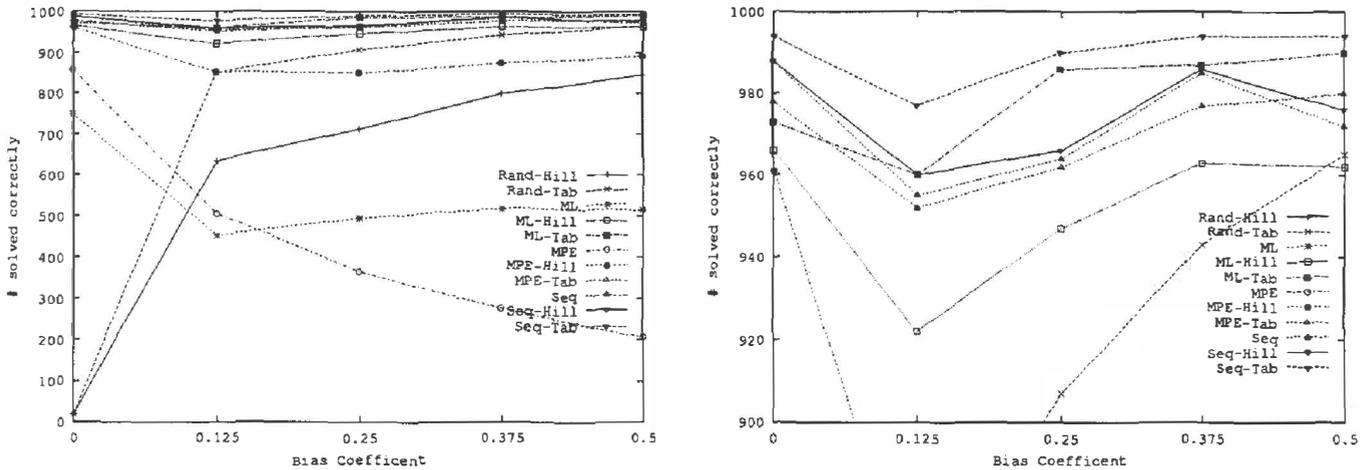

Figure 6: The solution quality of the various search and initialization methods for the second random generation method. The y-axis is the number of problems solved correctly out of 1000. The x-coordinate is the bias coefficient used for quantifying the CPTs. The plot on the right is a zoomed view of the one on the left.

Evaluations Required

| Method | Mean | Stdev | Max |
|---|---|---|---|
| Rand Hill | 12.5 | 2.5 | 21 |
| Rand Taboo | 14.3 | 11.0 | 144 |
| MPE | 1 | 0 | 1 |
| MPE Hill | 2.6 | 1.3 | 8 |
| MPE Taboo | 4.0 | 8.3 | 137 |
| ML | 1 | 0 | 1 |
| ML Hill | 1.6 | .74 | 4 |
| ML Taboo | 1.9 | 3.3 | 62 |
| Seq | 25 | 0 | 25 |
| Seq Hill | 25.0 | .04 | 26 |
| Seq Taboo | 25.0 | .9 | 45 |

Table 3: Statistics on the number of evaluations each method required before achieving the value it eventually returned. These are based on the random method 2, bias .5 data set. The statistics for the other data sets are similar.

over the first peak they discovered.[7] This suggests that one viable method for quick approximation is to simply climb to the first peak and return the result. Taboo search on the other hand was able to improve on the first peak in some cases.

---

[7]It appears that the random walk used in restarting does not make eventually selecting a better region very likely when using so few search steps. Often, when a sub optimal hill was encountered, the optimal hill was just 2 or 3 moves away. In those cases, the taboo search was usually able to find it (because its search was more guided), while random walking was not.

## 5 Discussion

The primary advantage of approximating MAP using local search in place of solving it exactly using structure-based methods is that local search typically requires much less time and space, yet produces very good approximations. Given a network with $n$ variables and an elimination order of width $w$, local search requires $O(n \exp(w))$ space. Standard exact algorithms require $O(n \exp(w_c))$ space, where $w_c$ is the width of a constrained elimination order. Moreover, the time complexity of local search is $O(in \exp(w))$, where $i$ is the number of search steps. An exact algorithm on the other, would require $O(n \exp(w_c))$ time. As our experiments have shown, $i$ can be quite small, while the difference between $\exp(w)$ and $\exp(w_c)$ can be quite significant. Therefore, many MAP problems that are intractable for exact methods can be approximated well and efficiently using local search.

Local search methods also have a big advantage over MPE and ML approximations (the methods typically used in place of MAP in diagnosis) in that it is much more accurate. With just a few (in some of our experiments 2-5) network evaluations, one can use ML or MPE to initialize, and then hill climb to produce a drastically better MAP solution. If more accuracy is desired, sequential initialization can be used with hill climbing or taboo search instead, at a cost of a few more network evaluations.

## References

[1] G. Cooper. Computational complexity of probabilistic inference using Bayesian belief networks

## A  Generating Random Networks

We generated several types of networks to perform our experiments. We used two methods for generating the structure, and a single parametric method for generating the quantification.

### A.1  Generating the Network Structure

The first method is parameterized by the number of variables $N$ and the connectivity $c$. This method tends to produce structures with widths that are close to $c$. See [6] for an algorithmic description.

The second method is parameterized by the number of variables $N$, and the probability $p$ of an edge being present. We generate an ordered list of $N$ variables, and add an edge between variables $X$ and $Y$ with probability $p$. The edges added are directed toward the variable that appears later in the order.

For the experiment in Figure 2, we used method 1 with $N = 100$ and $c$ between 1 and 20.

For the experiments in Figures 5 and 6, we used $N = 100$, $c$ between 6 and 12, and $p = .025$. These numbers were chosen so that the MAP width would be small enough that we could compute the exact value to measure the solution quality.

### A.2  Quantifying the Dependencies

The quantification method is parameterized by a bias parameter $b$. The values of the CPTs for the roots were chosen uniformly. The values for the rest of the nodes were based on a bias, where one of the values $v$ was chosen uniformly in $[0, b)$, and the other as $1 - v$. For example, for $b = .1$, each non root variable given its parents has one value in $[0, .1)$, and the other in $(.9, 1]$. Special cases $b = 0$, and $b = .5$ produce determistic, and uniformly random quantifications respectively.